\documentclass[12pt]{article}
\pdfoutput=1 

\usepackage{epsf} 
\usepackage{cite}
\usepackage{algorithmic}

\usepackage{amsmath}
\usepackage{amssymb}
\usepackage{graphicx}
\usepackage{breqn}
\usepackage{subcaption}
\usepackage{float}
\usepackage[multiple]{footmisc}
\restylefloat{table}
\usepackage[mathscr]{euscript}
\usepackage{bigints}
\usepackage[font={footnotesize}]{caption}
\usepackage{float}
\usepackage{color}
\usepackage[pdftex]{hyperref}
\usepackage{array}

\usepackage{graphicx}
\graphicspath{{./figs/}}

\begin{document}

\title{}

\title{\Large Clustering of Big Data with Mixed Features \footnote{Corresponding Author: Joshua Tobin (tobinjo@tcd.ie)}}
\author{Joshua Tobin\footnote{School of Computer Science \& Statistics, Trinity College Dublin.}
	\and Mimi Zhang\footnotemark[2]\textsuperscript{ ,}\footnote{I-Form Centre for Advanced Manufacturing, Trinity College Dublin, Ireland.}}

\maketitle

\begin{abstract}
Clustering large, mixed data is a central problem in data mining. Many approaches adopt the idea of $k$-means, and hence are sensitive to initialisation, detect only spherical clusters, and require a priori the unknown number of clusters. We here develop a new clustering algorithm for large data of mixed type, aiming at improving the applicability and efficiency of the peak-finding technique. The improvements are threefold: (1) the new algorithm is applicable to mixed data; (2) the algorithm is capable of detecting outliers and clusters of relatively lower density values; (3) the algorithm is competent at deciding the correct number of clusters. The computational complexity of the algorithm is greatly reduced by applying a fast $k$-nearest neighbors method and by scaling down to component sets. We present experimental results to verify that our algorithm works well in practice.\\
\textbf{Keywords}: Clustering; Big Data; Mixed Attribute; Density Peaks; Nearest-Neighbor Graph; Conductance.

\end{abstract}


\section{Introduction}
Partitioning large data of mixed type into homogeneous clusters is required in many data mining applications. A popular generalization of the $k$-means algorithm to mixed data is the $k$-prototypes method \cite{huangClusteringLargeData1997a}, in which the distance metric for categorical attributes is the 0-1 indicator function that indicates whether two categories match. The objective function can be optimized separately for the numerical and categorical parts, and the cluster-wise dominant category of each attribute will constitute the cluster center. However, the performance and applicability of the $k$-prototypes method is limited by the fact that (1) the distance metric does not distinguish between the different values taken by an attribute, (2) the cluster prototype provides no information on the within-cluster frequency distribution of each categorical attribute, and (3) the algorithm is sensitive to initialisation, requires a priori the number of clusters, and tends to detect spherical clusters of homogeneous size.

A broad range of remedies are available for problems (1) and (2). \cite{jiImprovedKprototypesClustering2013a} improve the $k$-prototypes method by modifying the distance metric and the cluster centers. The distance metric weights the Hamming distance by the frequency of each category in a cluster, and the cluster prototypes for each categorical attribute are defined as a list of tuples containing the attribute value and assigned weights equal to their relative frequency in the cluster. Another popular adaptation of the $k$-prototypes method is given by \cite{ahmadKmeanClusteringAlgorithm2007}. The difference between \cite{jiImprovedKprototypesClustering2013a} and \cite{ahmadKmeanClusteringAlgorithm2007} is that the latter defines another distance metric which requires more computations. A third $k$-prototypes method uses a semi-parametric approach to balance the contribution of numerical and categorical variables \cite{fossSemiparametricMethodClustering2016}. For problem (3), the works of \cite{jiNovelClusterCenter2015} and \cite{jinyinNovelClusterCenter2017} are aimed at reducing the impact of initialisation. In both works, quadratic complexity makes them impractical for large datasets. To determine the correct number of clusters, a typical approach is to apply a clustering validation index, e.g., \cite{liangDeterminingNumberClusters2012}; however, in terms of efficiency, no validation index comes out on top \cite{yaoImprovedClusteringAlgorithm2018}. When clusters are of varying size and density, $k$-means type algorithms are known to perform poorly \cite{xiongKMeansClusteringValidation2009a}.

Other than $k$-means type algorithms, hierarchical clustering was applied to mixed-type data by \cite{liUnsupervisedLearningMixed2002} and \cite{hsuMiningMixedData2007}, and model-based clustering methods were applied to mixed-type data by \cite{mcparlandModelBasedClustering2016}. A common drawback, however, is that they all have high computational complexity. The popular density-based clustering method by fast search and find of density peaks (DPC) \cite{Rodriguez1492} was adapted to mixed data in \cite{dingEntropybasedDensityPeaks2017}. To reduce the computational complexity in density calculation, one can compute a local estimate of the density by applying kernel functions on nearest neighbors; see \cite{Xie2016} and \cite{yaohuiAdaptiveDensityPeak2017}.

In DPC and it's extensions, cluster centers are commonly selected from the extreme points on the decision graph (gamma plot), a scatter plot of the products $\varphi \times \omega$, where $\varphi$ represents local density, and $\omega$ represents distance to the nearest neighbor of higher density. The remaining points are assigned to the same cluster as their nearest neighbor of higher density. However, in the decision graph, there is not always a distinguishing gap between true cluster centers and non-centers. \cite{liuClusteringSearchDescending2019} and \cite{linDensityPeakBasedClustering2017} present some analysis to the decision graph, which literately transforms the number of clusters to another subjective parameter. Automatic center-selection methods are developed in \cite{yaohuiAdaptiveDensityPeak2017} and \cite{wangMcDPCMulticenterDensity2020}, with the idea of initially selecting all candidate centers and then iteratively merging clusters; the complexity is of the order $O(n^2)$.

Motivated by the flexibility of the peak-finding technique, we here develop a new clustering algorithm for big data of mixed type. Our contributions are:
\begin{enumerate}
 \item We define a new distance metric that naturally balances the contributions from numerical attributes and categorical attributes. For each categorical attribute, the distance between any pair of categories incorporates the frequency distribution in the data.
 \item We utilize the concept of `component' from graph theory to make our algorithm capable of dealing with clusters of varying density.
 \item We utilize the concept of `conductance' from graph theory to automatically identify the correct number of clusters.
 \item Utilizing a fast $k$-nearest neighbors method, the complexity of our algorithm is of the order $O(n \log(n))$.
\end{enumerate}
The remainder of the paper contains three sections. Section \ref{sec2} includes all the methodological details. Section \ref{PA} defines the distance metrics and explains the weighting mechanism.
 Section \ref{PB} presents the limitations of the DPC method which then motivate the CPF method. Section \ref{PC} lists the steps of the CPF algorithm, and Section \ref{PD} details the center-selection part. The excellent performance of the CPF method is demonstrated in Section \ref{sec4}. The work concludes in Section \ref{sec5} with a summary of the contributions of the paper and discussion of future avenues of research.

\section{The Clustering Method}\label{sec2}
\subsection{The Distance Metric}\label{PA}
Let $\mathbb{X}$ denote the set of available data: $\mathbb{X}=\{\pmb{x}_1, \ldots, \pmb{x}_n\}$, where each sample $\pmb{x}_i$ ($i=1, \ldots, n$) is described by $(p_1+p_2)$ features. The first $p_1 (>0)$ features are categorical, while the remaining $p_2 (>0)$ features are numerical (including ordinal). We might assume that the samples of each numerical feature have been standardized to have mean zero and unit variance. (To eliminate the influence of outliers, the standardization only involves 98\% of the data.) Let $A_j=\{a_j^q: q=1, \ldots, |A_j|\}$ denote the set of all possible categories for the categorical feature $X_j$ ($1\leq j\leq p_1$). To define an appropriate distance metric, we take the dummy-variable approach in which the categorical feature $X_j$ is replaced by a vector of $|A_j|$ binary variables: $\pmb{b}^j\in\{0, 1\}^{|A_j|}$. The samples of the categorical feature $X_j$ are then replaced by $\{\pmb{b}^j_1, \ldots, \pmb{b}^j_n\}$. For notational convenience, we might let $\pmb{z}\in\mathbb{R}^{p_2}$ denote the vector of the $p_2$ numerical features, and $\mathbb{Z}=\{\pmb{z}_1, \ldots, \pmb{z}_n\}$ denote the set of numerical data.

Let $\pmb{v}$ denote the dummy counterpart of an arbitrary point $\pmb{x}$, where $\pmb{v}$ is concatenated by $\{\pmb{v}^1\in\{0, 1\}^{|A_1|}, \ldots, \pmb{v}^{p_1}\in\{0, 1\}^{|A_{p_1}|}, \pmb{v}^0\in\mathbb{R}^{p_2}\}$. Let $\circ$ denote the element-wise multiplication. The distance between the two points $\pmb{x}_i$ and $\pmb{x}$ is defined as follows:
\begin{equation}
\label{NewD}
d(\pmb{x}_i, \pmb{x})^2=\sum_{j=1}^{p_1}\rho_j\|\sqrt{\pmb{w}^j}\circ(\pmb{b}^j_i-\pmb{v}^j)\|_2^2+\|\pmb{z}_i-\pmb{v}^0\|_2^2,
\end{equation}
where $\|\cdot\|_2$ is the Euclidean norm. The two types of weights, $\{\rho_1, \ldots, \rho_{p_1}\}$ and $\{\pmb{w}^1, \ldots, \pmb{w}^{p_1}\}$, are explained as follows.

The weight vector $\pmb{w}^j\in \mathbb{R}_{+}^{|A_j|}$ consists of the weights for the categories in $A_j$. We here propose two weighting mechanisms. If one wants to give large weights to frequent categories, $\pmb{w}^j$ can be defined as the relative frequency vector $\frac{1}{n}\sum_{i=1}^{n}\pmb{b}^j_i$. If one wants to give large weights to rare categories, $\pmb{w}^j$ can be defined as the normalized value of $-\log(\frac{1}{n}\sum_{i=1}^{n}\pmb{b}^j_i)$. The distance metric in (\ref{NewD}) can be readily applied to center-finding clustering algorithms: for any cluster $\mathbb{C}$, the cluster center is simply the sample mean: $\frac{1}{|\mathbb{C}|}\sum_{\pmb{x}_i\in\mathbb{C}}\pmb{z}_i$ for the numerical part and $\frac{1}{|\mathbb{C}|}\sum_{\pmb{x}_i\in\mathbb{C}}\pmb{b}^j_i$ for the categorical feature $X_j$ ($1\leq j\leq p_1$).

The feature weights $\{\rho_1, \ldots, \rho_{p_1}\}$ are to make the categorical features comparable to the numerical features. The contribution from the numerical features to the squared distance $d(\pmb{x}_i, \pmb{x})^2$ is $\|\pmb{z}_i-\pmb{v}^0\|_2^2$. With the samples of each numerical feature standardized to have unit variance, we have $\mbox{E}\|\pmb{z}_i-\pmb{v}^0\|_2^2=2p_2$; that is, the expected squared difference between two random samples of a numerical feature is 2. Therefore, to make the expected contribution from a categorical feature equal to the expected contribution from a numerical feature, the weight $\rho_j$ need satisfy the constraint $\mbox{E}[\rho_j\|\sqrt{\pmb{w}^j}\circ(\pmb{b}^j_i-\pmb{v}^j)\|_2^2]=2$. Let $\{\hat{p}_j^q: q=1, \ldots, |A_j|\}$ denote the sample proportions for all the categories of the categorical feature $X_j$ ($1\leq j\leq p_1$). We have
\begin{displaymath}
\mbox{E}\|\sqrt{\pmb{w}^j}\circ(\pmb{b}^j_i-\pmb{v}^j)\|_2^2=\sum_{q=1}^{|A_j|}\sum_{\tilde{q}\neq q}(w_q^j +w_{\tilde{q}}^j )\hat{p}_j^q\hat{p}_j^{\tilde{q}}.
\end{displaymath}
Hence, the weight $\rho_j$ is calculated from
\begin{displaymath}
\rho_j=\frac{2}{\sum_{q=1}^{|A_j|}\sum_{\tilde{q}\neq q}(w_q^j +w_{\tilde{q}}^j )\hat{p}_j^q\hat{p}_j^{\tilde{q}}}.
\end{displaymath}
If $\pmb{w}^j$ is a vector of 1's, we have $\rho_j=\frac{1}{\sum_{q=1}^{|A_j|}\hat{p}_j^q(1-\hat{p}_j^q)}.$ The denominator is the Gini index, which measures the uncertainty/variance of the categorical feature $X_j$. Therefore, we can interpret the feature weight $\rho_j$ as standardizing the categorical feature $X_j$ to have unit variance.

\subsection{The Motivations}\label{PB}
The peak-finding method \cite{Rodriguez1492} is built on the assumptions that (1) a cluster center has a relatively higher local density value than its neighbors, and (2) a cluster center is distant from any point that has a higher local density value. Two indices are calculated for each data point $\pmb{x}_i$: its local density $\varphi_i$ and the minimum of the distances from points that have higher local density values
\begin{equation}\label{distance}
\omega_i= \begin{cases}
             \max\{d(\pmb{x}_i, \pmb{x}_j): 1\leq j\leq n\},\parbox[t]{.3\columnwidth}{if $\varphi_i$ is the largest;} \\
             \min\{d(\pmb{x}_i, \pmb{x}_j): 1\leq j\leq n, ~\varphi_j>\varphi_i\}, \hbox{otherwise.}
           \end{cases}
\end{equation}
Here, $d(\pmb{x}_i, \pmb{x}_j)$ is the distance between the data points $\pmb{x}_i$ and $\pmb{x}_j$. $\omega_i$ will be large if $\varphi_i$ is a local or global maximum, or if $\pmb{x}_i$ is an outlier. Therefore, a data point $\pmb{x}_i$ will be treated as a cluster center, only if both $\varphi_i$ and $\omega_i$ are large.

\begin{figure}[!h]
 \centering
 \includegraphics[width=\linewidth]{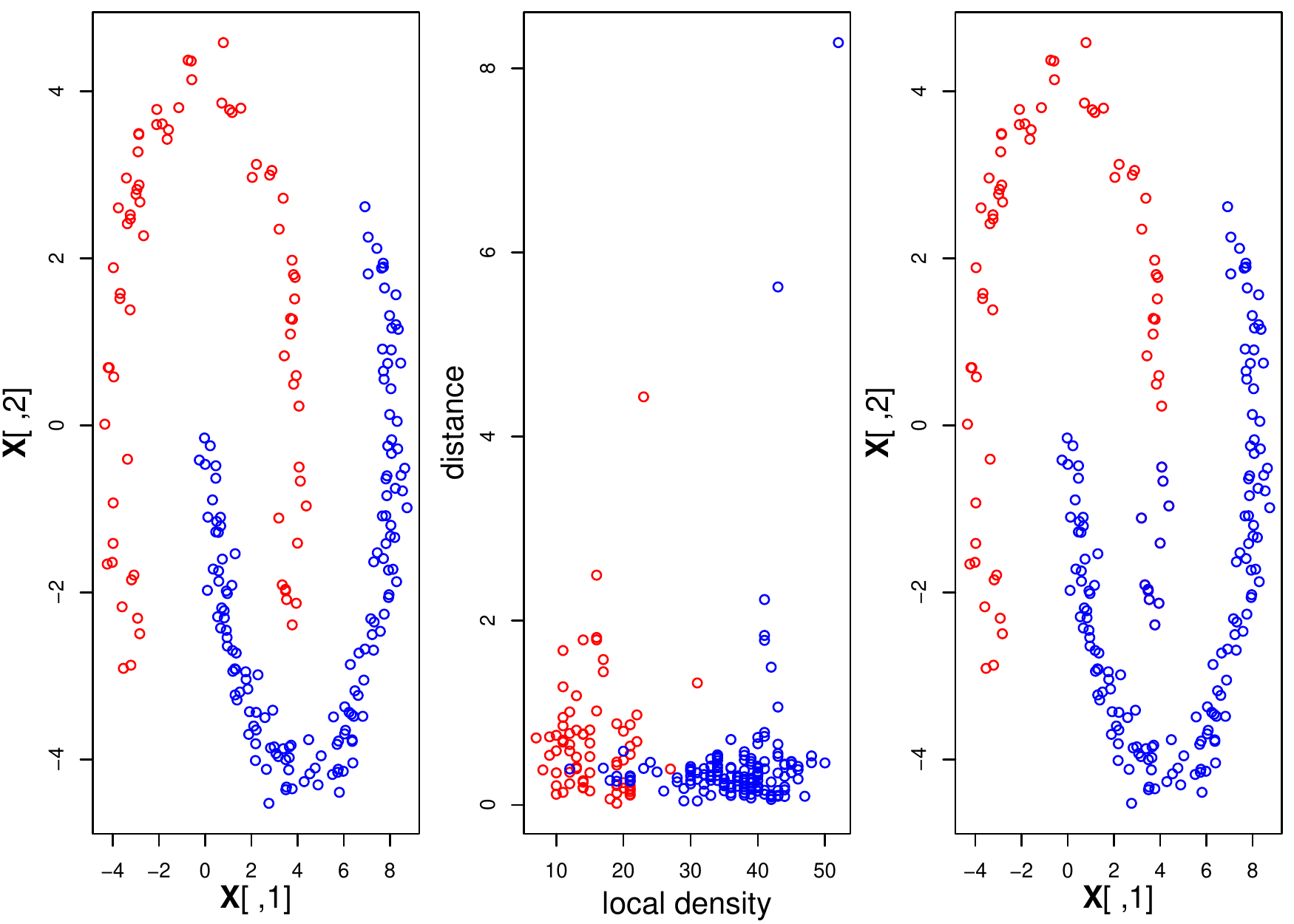}\\
 \caption{A simulated example showing certain drawbacks of the original peak-finding method. Left panel: data are from two clusters with different degrees of denseness. Middle panel: the decision graph can not identify the center for the red (sparse) cluster. Right: though given the true cluster centers, some samples are still incorrectly assigned.}
 \label{problems}
\end{figure}

We here apply the original peak-finding method to simulated data with two numerical features to explain its drawbacks and hence our motivations. The data from two clusters are plotted in the left panel of Figure \ref{problems},
with the red cluster containing 75 samples, and the blue cluster containing 150 samples. It is clear that the data in the blue cluster are denser than those in the red cluster.
\begin{enumerate}
  \item The middle panel is the decision graph, where the local density and distance are calculated via Equations (1) and (2) in \cite{Rodriguez1492}. Samples in red are from the red cluster, and samples in blue are from the blue cluster. The middle panel shows that most of the data points in the blue cluster have a larger local density value than those in the red cluster. If we only select two data points as the cluster centers, we will certainly pick the two blue points in the top right corner, which means that the data points in the red cluster will all be incorrectly grouped.
  \item Let the top right blue point and the top right red point in the decision graph be the two cluster centers. The right panel shows the final clustering, adopting the assignment strategy given in \cite{Rodriguez1492}. The figure implies that, even if we know the true cluster centers, the final clustering may still be wrong. In \cite{Rodriguez1492}, after the cluster centers are identified, each remaining point is assigned to the same cluster as its nearest neighbor of higher density (not directly assigned to its closest cluster center). However, as indicated by the right panel, the distance from a red point $\pmb{x}_i$ to its nearest red point of higher density is larger than the distance from $\pmb{x}_i$ to its nearest blue point of higher density. Hence the red point is assigned to the blue cluster. If one red point is incorrectly assigned, then the following assignments of its neighboring red points are highly likely to be wrong.
  \item In \cite{Rodriguez1492}, selecting cluster centers from the decision graph is a subjective task which is difficult to do correctly. According to the middle panel, it is likely that one will mistakenly select two blue samples and one red sample as the cluster centers. We need an algorithm to automatically return the correct number of cluster centers; otherwise, the requisite statistical knowledge in selecting cluster centers will impede the wide application of the peak-finding technique.
\end{enumerate}

\subsection{The CPF Method}\label{PC}
To address problems (1) and (2), we here introduce the concept of \textit{component} from graph theory. Let $G=(\mathbb{X}, E)$ be an undirected and unweighted $k$-nearest neighbor graph; the vertex set is the data set $\mathbb{X}$, and the edge set is $E$. There is an edge between two vertices $\pmb{x}_i$ and $\pmb{x}_j$, denoted by $\{\pmb{x}_i, \pmb{x}_j \}\in E$, if and only if $\pmb{x}_i$ and $\pmb{x}_j$ are among the $k$ nearest neighbors of each other. 
Note that, if $\pmb{x}_j$ is a $k$-nearest neighbor of $\pmb{x}_i$, but $\pmb{x}_i$ is not a $k$-nearest neighbor of $\pmb{x}_j$, then $\{ \pmb{x}_i, \pmb{x}_j \}\notin E$. Particularly, if $\pmb{x}_i$ is an outlier, and all its $k$-nearest neighbors are located within dense clusters, then there will be no edge connected to $\pmb{x}_i$. Hence, we can claim that vertices in the graph $G$ with very few or no edges are outliers.

A path of length $m$ from $\pmb{x}_i$ to $\pmb{x}_j$ is a sequence of distinct edges in $E$, starting at vertex $v_0= \pmb{x}_i$ and ending at vertex $v_m = \pmb{x}_j$, $\{\pmb{x}_i, v_1\}, \{v_1, v_2\}, \ldots, \{v_{m-1}, \pmb{x}_j \}$ such that $\{v_{r-1}, v_r \} \in E$ for all $r=1, \ldots, m$. We say that the two data points $\pmb{x}_i$ and $\pmb{x}_j$ are connected, if there is a path from $\pmb{x}_i$ to $\pmb{x}_j$ in the graph $G$. A component of $G$ is a subgraph in which any two vertices are connected to each other by paths. From the definition of component, we know that the components of $G$ reveal certain underlying pattern of the data. In particular, the data can be partitioned into disjoint subsets, herein called component sets. Intuitively, two data points belonging to two different component sets are highly likely to belong to different clusters. Therefore, we propose to apply the peak-finding method on each individual component set, not on the whole dataset $\mathbb{X}$, to find at least one cluster center for each component set.

While samples in different component sets belong to different clusters, samples within one component set could likewise belong to different clusters -- when the component set contains multiple clusters. The CPF method is detailed as follows:
\begin{enumerate}
  \item Given the value $k$,  create an undirected and unweighted $k$-nearest neighbor graph $G=(\mathbb{X}, E)$.
  \item Let $\mathbb{O}_1$ denote the vertices/samples that have no edge. Partition the data $\mathbb{X}\setminus\mathbb{O}_1$ into disjoint component sets according to the graph $G$.
  \item For each component set $\mathbb{C}\subseteq\mathbb{X}\setminus\mathbb{O}_1$,
  \begin{enumerate}
    \item for any sample $\pmb{x}_i\in\mathbb{C}$, find its $K$ nearest neighbors $\mathscr{N}_K(\pmb{x}_i)\subseteq\mathbb{C}$ and calculate the local density $\varphi_i=\sum_{\pmb{x}\in\mathscr{N}_K(\pmb{x}_i)}\exp(-d(\pmb{x}_i, \pmb{x}))$;
    \item for any sample $\pmb{x}_i\in\mathbb{C}$, calculate the distance $\omega_i$:
		\begin{equation*}
		\omega_i=
		           \begin{cases}
		             \max\{d(\pmb{x}_i, \pmb{x}_j): \pmb{x}_j\in\mathbb{C}\}, \parbox[t]{.25\columnwidth}{if $\varphi_i$ is the largest in $\mathbb{C}$;} \\

		             \min\{d(\pmb{x}_i, \pmb{x}_j): \pmb{x}_j\in\mathbb{C}, ~\varphi_j>\varphi_i\}, \\\parbox[t]{.23\columnwidth}{otherwise}\\

		          \end{cases}
		\end{equation*}
\item plot the decision graph $\{(\varphi_i, \omega_i): \pmb{x}_i\in\mathbb{C}\}$.
 \end{enumerate}
  \item According to the decision graphs, select the outliers, denoted by $\mathbb{O}_2$, and select the cluster centers for each component set.
  \item For a non-center point $\pmb{x}_i\in\mathbb{C}\setminus\mathbb{O}_2$, find the nearest neighbor of higher density $\hat{\pmb{x}}\in\mathbb{C}$. Assign $\pmb{x}_i$ to the same cluster as $\hat{\pmb{x}}$.
\end{enumerate}

The local density defined in step $3 (a)$ is different from the original: $\sum_{j=1}^{n}\delta(d(\pmb{x}_i, \pmb{x}_j)<d_c)$, where $\delta(\cdot)$ is the indicator function, and $d_c$ is an input parameter. While the two definitions are of the same power in decision making, the parameter $K$ is much easier to tune than the cutoff parameter $d_c$.

The $K$ value for calculating the local density $\varphi_i$ is the same for all the component sets, e.g., $K=\sqrt{n}$. However, if the size of a component set $\mathbb{C}$ is smaller than $K$, there will be fewer than $K$ neighbors for any $\pmb{x}_i \in \mathbb{C}$. Hence, for a small component set $\mathbb{C}$, we let $K$ take the value $|\mathbb{C}|$. Moreover, to reduce the computational load when dealing with big data, one can put an upper limit on $K$, e.g., $K\leq100$.

\subsection{Center Selection}\label{PD}
To address problem (3), we utilize the idea of \textit{conductance} from graph theory. Let $G(\mathbb{C}) = (\mathbb{C}, E(\mathbb{C}))$ be the subgraph of $G = (\mathbb{X}, E)$, where $\mathbb{C}$ is a component set, and $E(\mathbb{C})$ is the induced edge set. We further assign weights to the edges in $G(\mathbb{C})$: the weight for the edge $\{\pmb{x}_i, \pmb{x}_j \}\in E(\mathbb{C})$ is $w(\{\pmb{x}_i, \pmb{x}_j \}) = \exp(-d(\pmb{x}_i, \pmb{x}_j))$; that is, the weight for an edge is inversely related to the distance between the two vertices. For a subset $S \subseteq \mathbb{C}$, we denote by $w(S) = \sum_{\pmb{x}_i \in S} \sum_{\pmb{x}_j \in \mathbb{C}} w (\{ \pmb{x}_i, \pmb{x}_j \})$, the total weight of edges connected to vertices in $S$. A cut on the graph $G(\mathbb{C})$ is a partition of $\mathbb{C}$ into two nonempty subsets $S$ and $\bar{S}$ such that $S \cap \bar{S} = \varnothing$ and $S \cup \bar{S} = \mathbb{C}$. The conductance of a cut $(S, \bar{S})$ of $\mathbb{C}$ is defined as:
\begin{equation}
\Phi(S, \bar{S}; G(\mathbb{C})) = \frac{\sum_{ \pmb{x}_i \in S,  \pmb{x}_j \in \bar{S}} w (\{ \pmb{x}_i, \pmb{x}_j \} )}{\min \{ w(S), w(\bar{S}) \}}.
\end{equation}

Conductance is a measure of the quality of the cut $(S, \bar{S})$. Conductance captures the notion that clusters should be sets of objects with stronger intra-cluster connections than inter-cluster connections \cite{leskovecStatisticalPropertiesCommunity2008}. A high-quality cut will produce a low conductance value. Conductance has been widely used for graph clustering applications \cite{schaefferGraphClustering2007, gleichVertexNeighborhoodsLow2012}. Our center selection method is motivated by the insight from \cite{leskovecStatisticalPropertiesCommunity2008} that, when comparing clusterings of varying cluster numbers, local minima of the conductance values corresponds to the best clustering.

\begin{figure}[!h]
	\centering
	\includegraphics[width=0.8\linewidth]{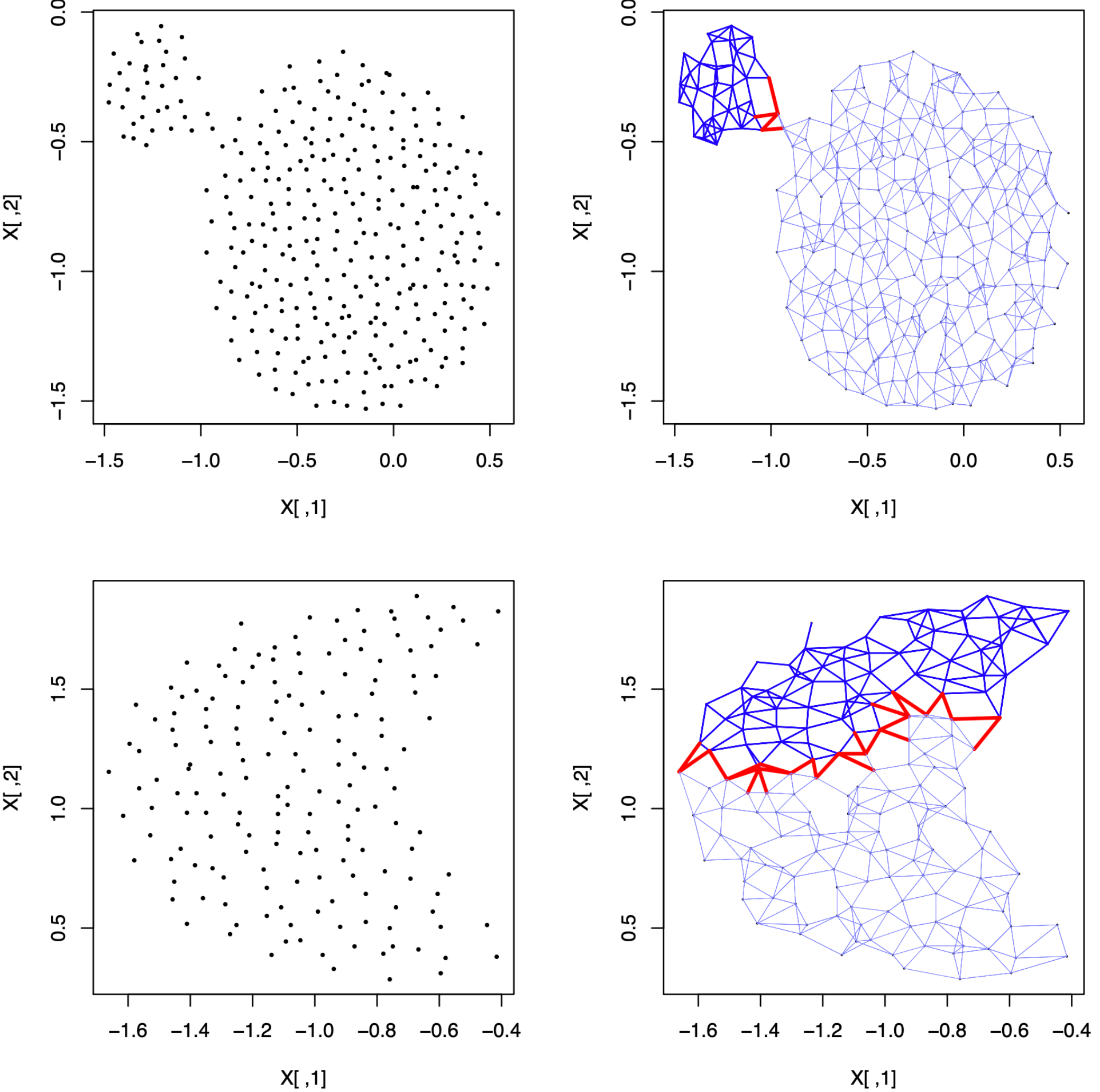}\\
	\caption{Left panels: data from two component sets; the upper contains two clusters, and the lower consists of one cluster. Right panels: cuts on the subgraphs; inter-cluster edges are in red, and inter-cluster edges are in blue. The conductance value of the upper cut is smaller than the conductance value of the lower cut.}
	\label{conduc}
\end{figure}

As an illustrative example, the left panels in Figure \ref{conduc}
show the scatter plots of two component sets (left top and left bottom). The top component set consists of two clusters, while the samples in the bottom component set belong to one single cluster. The right panels show the clusterings when we select two cluster centers for both component sets. The cuts are indicated by the red edges that connect two clusters. Figure \ref{conduc} implies that the conductance value of a clear cut will be different from the conductance value of a redundant cut.

The gamma values $\{\gamma(\pmb{x}_i)=\varphi_i\times\omega_i: \pmb{x}_i\in\mathbb{C}\}$ will automatically decide the candidate cluster centers, denoted by $\{\pmb{x}_{(1)}, \pmb{x}_{(2)}, \ldots, \pmb{x}_{(\beta)}\}$, where the candidates are ordered according to their gamma values. However, $\beta$ is generally larger than the true number of clusters. Here, $\pmb{x}_{(1)}$ is the sample having the largest gamma value and therefore is directly selected as the cluster center. Our algorithm to decide whether to select more samples as the cluster centers proceeds as follows.
\begin{enumerate}
  \item Select $\pmb{x}_{(2)}$ as the candidate cluster center and partition the component set $\mathbb{C}$ into two disjoint clusters, denoted by $S_1$ and $S_2$, where non-center samples are assigned to the same cluster as their nearest neighbor of higher density. Calculate the conductance $\Phi(S_1, S_2; G(\mathbb{C}))$.
      \begin{enumerate}
        \item Determine the minimal neighborhood value $k$, denoted by $\tilde{k}$, such that the subgraphs $G_{\tilde{k}}(S_1)$ and $G_{\tilde{k}}(S_2)$ of the  $k$-nearest neighbor graph $G_{\tilde{k}}(\mathbb{C})$ are connected.
        \item Calculate the conductance $\Phi(S_1, S_2; G_{\tilde{k}}(\mathbb{C}))$.
        \item Set $\hat{k}=\tilde{k}+1$, create the new $k$-nearest neighbor graph $G_{\hat{k}}(\mathbb{C})$, and calculate the conductance $\Phi(S_1, S_2; G_{\hat{k}}(\mathbb{C}))$.
       \item If $\Phi(S_1, S_2; G_{\hat{k}}(\mathbb{C}))>\Phi(S_1, S_2; G_{\tilde{k}}(\mathbb{C}))$, then $\pmb{x}_{(2)}$ is not a cluster center, and the algorithm terminates. Otherwise, $\pmb{x}_{(2)}$ is the cluster center, and define $\Phi_2=\Phi(S_1, S_2; G(\mathbb{C}))$.
      \end{enumerate}
   \item For $j=3, \ldots, \beta$,
  \begin{enumerate}
   \item given the cluster centers $\{\pmb{x}_{(1)}, \pmb{x}_{(2)}, \ldots, \pmb{x}_{(j)}\}$, partition the component set $\mathbb{C}$ into disjoint clusters $\{S_1, S_2, \ldots, S_j\}$;
    \item calculate the conductance values: $\Phi(S_t,  \bar{S}_t; G(\mathbb{C}))$, for $1\leq t\leq j$, where $\bar{S}_t=\mathbb{C}\setminus S_t$;
    \item define $\Phi_j=\max\{\Phi(S_t, \bar{S}_t; G(\mathbb{C})): 1\leq t\leq j\}$.
  \end{enumerate}
  \item The true cluster centers are $\{\pmb{x}_{(1)}, \pmb{x}_{(2)}, \ldots, \pmb{x}_{(\beta^*)}\}$, where $\beta^*=\arg\min_{2\leq j\leq\beta} \{\Phi_2, \Phi_3, \ldots, \Phi_\beta\}$.
\end{enumerate}

Conductance is a powerful index for identifying small or sparse clusters, as the sum of inter-cluster edge weights is scaled by the smaller cluster. Consequently, the CPF method is effective at detecting clusters of arbitrary shape, size and density.

The most computation-intensive task is creating the graph $G(\mathbb{X}, E)$, which requires $O(n \log(n))$ operations utilizing the method developed by \cite{zhangFastKNNGraph2013a}. Another major computational burden is finding for each non-center point the nearest neighbor of higher density. The majority of points have an instance of higher density in their $k$ nearest neighbors. For the remaining points, a broad search must be undertaken. The search is conducted over all the instances in the component.

\section{Numerical Study}\label{sec4}
\subsection{Experimental Set Up}
To evaluate the CPF method, we adopt five widely used external indices: the {Adjusted Rand Index} (ARI), the {Purity Score} (PS), the {F1-Score} (F1), the {Normalized Mutual Information} (NMI), and the {Clustering Accuracy} (CA). For each of these metrics, a larger value indicates a higher-quality clustering. Due to space limitations, the formal definitions of these metrics are omitted, and can be found in \cite{xiongKMeansClusteringValidation2009a, rahmanHybridClusteringTechnique2014}.

Given the characteristics of the CPF method, we evaluate its performance by (1) applying it to numerical, categorical and mixed-attribute datasets of medium or large size and comparing it with other representative clustering methods, and  (2) applying it to large datasets and analyzing the run time of CPF.

25 datasets collected from the UCI machine learning repository are used for the experiments. For the first part of our analysis, 22 datasets are used, nine with numerical attributes only, four with categorical attributes only, and nine with mixed attributes. Their details are summarized in Table \ref{TableDS1}.
\begin{table}
	\tiny
	\centering
	\begin{tabular}{c c c c }
		\hline
		Dataset & Instances & Features(Num.-Cat.) & Classes \\
		\hline
		Dermatology&	358	&34 (34-0)&	6\\
		Ecoli &	336	& 7 (7-0)&	8\\
		Glass&	214&	9 (9-0)	&7\\
		Magic&	19020	&10 (10-0)&	2\\
		Mamm. Masses&830&	5 (5-0)	&2\\
		Page-Blocks&	5473&	10 (10-0)&	5\\
		Transfusion&	748	& 4 (4-0)&	2\\
		Wine Quality&4898&	11 (11-0)	&7\\
		Yeast&1484&	8 (8-0)	&10\\
		Kr vs. Kp&	3196&	36 (0-36)&	2\\
		Tic Tac Toe&	958	& 9 (0-9)&	2\\
		Mushroom&	8124& 	22 (0-22)&	2\\
		Breast Cancer&	277	& 9 (0-9)&	2\\
		German&	1000&	20 (7-13)&	2\\
		Credit&	653&	15 (6-9)	&2\\
		Adult&	30162&	14 (6-8)	&2\\
		CMC	&1473&	9 (1-8)	&3\\
		KDD '99 RI vs. B&	2225&	33 (29-4&	2\\
		KDD '99 BO vs. B&	2233	& 27 (25-2)&	2\\
		KDD '99 L vs. S&15913&	33 (27-6)&	2\\
		KDD '99 L vs. P&	10434&28 (23-5)&	2\\
		KDD '99 GP vs. S&	15945	&32 (27-5)&2 \\
		\hline
	\end{tabular}
	\caption{	\label{TableDS1} Characteristics of datasets for first experimental section.}
\end{table}
 The KDD '99 datasets are created by extracting two classes from the KDD Cup '99 competition dataset. To demonstrate the performance of CPF on large datasets, we use three mixed attribute datasets. The details of these datasets are summarized in Table \ref{TableDS2}.
\begin{table}
	\tiny
	\centering
	\begin{tabular}{c c c c }
		\hline
		Dataset & Instances & Features(Num.-Cat.) & Classes \\
		\hline
		Cov Type&581012&	12 (10-2)&	7\\
		Poker&1000000&10 (5-5)&	2\\
		KDD '99 DOS vs. NORM& 4898430 &41 (33-8)&2 \\
		\hline
	\end{tabular}
	\caption{	\label{TableDS2} Characteristics of datasets for second experimental section.}
\end{table}
All instances with missing values and all features taking only one value have been removed.

There are  three tuning parameters in the CPF method: the weighting scheme in the distance metric, the number of nearest neighbors used to construct the $k$-nearest neighbor graph ($k$), and the number of neighbors used to compute the local density ($K$). In what follows, the weighting scheme which uses the relative frequency vector is denoted by W1, and the scheme which uses the normalized log value is denoted by W2. Following \cite{maierOptimalConstructionKnearestneighbor2009}, we repeat the CPF procedure for values of $k$ ranging from 3 to 75. The values of the parameter $K$ are varied in intervals of 5 from 10 to 150. For each dataset, we report the results from the best combination of the three tuning parameters. The parameter $\beta$ will not affect the clustering if it is set to a value larger than the true number of clusters. As such, $\beta$ is set to $50$ for each dataset in the analysis.

\subsection{Clustering Numerical, Categorical \& Mixed Datasets}\label{ncm}
In this section, we conduct experiments to investigate the ability of CPF to cluster the datasets in Table \ref{TableDS1}. For the numerical datasets, the results are compared with the clusterings generated by $k$-means, GenClust++ \cite{islamCombiningKMeansGenetic2018}, and AD $k$-means \cite{ahmadKmeanClusteringAlgorithm2007}. For categorical datasets, $k$-modes \cite{huangExtensionsKMeansAlgorithm1998}, GenClust++ and AD $k$-means are used. $k$-prototypes \cite{huangClusteringLargeData1997a}, GenClust++, AD $k$-means, and KAMILA \cite{fossSemiparametricMethodClustering2016} are applied to the mixed attribute datasets. Source code was downloaded for each of the methods. Parameters were specified as suggested in each work. For CPF and GenClust++, the number of clusters is not required as an input parameter. For the remaining methods, the true number of classes is used as an input. For all methods bar CPF and GenClust++, the mean results for ten independent runs is presented.

Table \ref{TableRN}
\begin{table}[!h]
	\tiny
	\centering
\begin{tabular}{cccccc}
\hline
& &               &Gen       &AD           &      \\
& &k-means&Clust++&k-means&CPF\\
\hline
&ARI&0.701&0.761&0.001&\textbf{0.845}\\
&PS&0.858&\textbf{0.989}&0.310&0.916\\
Dermatology&F1&0.241&0.003&0.159&\textbf{0.304}\\
&NMI&0.849&0.842&0.010&\textbf{0.873}\\
&CA&0.196&0.003&0.165&\textbf{0.304}\\
\hline
&ARI&\textbf{0.517}&0.438&0.460&0.513\\
&PS&0.821&\textbf{0.946}&0.783&0.740\\
Ecoli&F1&0.343&\textbf{0.522}&0.114&0.019\\
&NMI&\textbf{0.626}&0.500&0.564&0.577\\
&CA&0.283&\textbf{0.432}&0.101&0.404\\
\hline
&ARI&0.209&0.016&0.228&\textbf{0.245}\\
&PS&0.542&\textbf{0.986}&0.539&0.575\\
Glass&F1&0.013&\textbf{0.491}&0.191&0.254\\
&NMI&0.304&0.039&0.326&\textbf{0.382}\\
&CA&0.056&\textbf{0.327}&0.203&0.254\\
\hline
&ARI&0.013&0.000&0.000&\textbf{0.026}\\
&PS&0.648&0.510&0.648&\textbf{0.709}\\
Magic&F1&0.288&0.478&\textbf{0.509}&0.000\\
&NMI&0.003&0.000&0.000&\textbf{0.032}\\
&CA&0.227&0.490&\textbf{0.497}&0.000\\
\hline
&ARI&0.000&\textbf{0.317}&-0.001&0.053\\
&PS&0.516&0.782&0.514&\textbf{0.849}\\
Mamm.&F1&0.352&\textbf{0.480}&0.238&0.053\\
Masses&NMI&0.000&\textbf{0.258}&-0.001&0.141\\
&CA&\textbf{0.516}&0.218&0.244&0.050\\
\hline
&ARI&0.101&0.002&0.013&\textbf{0.386}\\
&PS&0.905&0.450&0.898&\textbf{0.960}\\
Page&F1&0.010&\textbf{0.309}&0.201&0.001\\
Blocks&NMI&0.076&0.031&0.006&\textbf{0.273}\\
&CA&0.049&\textbf{0.420}&0.152&0.001\\
\hline
&ARI&0.065&0.021&0.021&\textbf{0.072}\\
&PS&0.762&0.580&0.761&\textbf{0.825}\\
Transfusion&F1&\textbf{0.694}&0.381&0.078&0.110\\
&NMI&0.011&0.009&0.002&\textbf{0.028}\\
&CA&\textbf{0.717}&0.420&0.140&0.110\\
\hline
&ARI&0.034&0.000&0.001&\textbf{0.139}\\
&PS&0.477&\textbf{1.000}&0.449&0.559\\
Wine&F1&0.155&0.064&\textbf{0.188}&0.139\\
Quality&NMI&0.069&0.000&0.002&\textbf{0.076}\\
&CA&0.118&0.033&\textbf{0.146}&0.029\\
\hline
&ARI&0.034&-0.001&0.012&\textbf{0.041}\\
&PS&0.477&\textbf{0.998}&0.352&0.719\\
Yeast&F1&0.029&\textbf{0.282}&0.123&0.004\\
&NMI&0.069&0.001&0.015&\textbf{0.174}\\
&CA&0.118&\textbf{0.164}&0.105&0.004\\
\hline
\end{tabular}
\caption{\label{TableRN} Results for datasets with numerical attributes. For each case, the winner is highlighted in bold.}
\end{table}
reports the results for numerical-attribute datasets. CPF outperforms the benchmark methods across the variety of metrics used. The design of the CPF method is validated by these results. Construction of the connected components reveals the underlying class structure for the Dermatology, Glass and Magic datasets. CPF performs well for these methods. For the Transfusion and Wine Quality datasets, the peak-finding method used in CPF allows for the detection of non-spherical clusters, and clusters of varying sizes. This contrasts positively with the benchmark methods which tend to return spherical clusters of homogeneous size.

Results for categorical-attribute datasets are reported in Table \ref{TableCN}
\begin{table}[!t]
	\tiny
	\centering
\begin{tabular}{cccccc}
\hline
&  &                 & Gen       & AD            &        \\
&  & $k$-modes& Clust++ &  $k$-means & CPF\\
\hline
&ARI&-0.040&0.005&0.021&\textbf{0.034}\\
&PS&0.717&0.047&0.717&\textbf{0.843}\\
Krvs.Kp&F1&0.041&0.000&\textbf{0.484}&0.027\\
&NMI&0.028&0.061&0.026&\textbf{0.082}\\
&CA&0.040&0.011&\textbf{0.165}&0.027\\
\hline
&ARI&0.011&0.007&0.028&\textbf{0.612}\\
&PS&0.653&0.040&0.655&\textbf{1.000}\\
TicTacToe&F1&0.148&0.002&\textbf{0.501}&0.015\\
&NMI&0.004&0.091&0.014&\textbf{0.596}\\
&CA&0.164&0.023&\textbf{0.491}&0.015\\
\hline
&ARI&\textbf{0.623}&0.035&0.118&0.012\\
&PS&0.895&0.072&0.661&\textbf{0.959}\\
Mushroom&F1&0.004&0.003&\textbf{0.503}&0.006\\
&NMI&\textbf{0.584}&0.274&0.101&0.194\\
&CA&0.003&0.029&\textbf{0.527}&0.006\\
\hline
&ARI&-0.001&0.001&0.140&\textbf{0.207}\\
&PS&0.776&0.166&0.775&\textbf{0.780}\\
Breast&F1&0.442&0.013&\textbf{0.510}&0.083\\
Cancer&NMI&-0.003&0.026&0.055&\textbf{0.079}\\
&CA&0.350&0.051&\textbf{0.500}&0.083\\
\hline
\end{tabular}
\caption{\label{TableCN}	Results for datasets with categorical attributes. For each case, the winner is highlighted in bold.}
\end{table}
and for mixed-attribute datasets in Table \ref{TableMN}.
\begin{table}[h!]
	\tiny
	\centering
\begin{tabular}{ ccccccc}
\hline
&  &                &       Gen &           AD &                 &        \\
&  & $k$-proto.& Clust++ & $k$-means & KAMILA & CPF\\
\hline
&ARI&0.027&0.005&0.006&0.040&\textbf{0.048}\\
&PS&0.700&0.051&0.700&0.700&\textbf{0.740}\\
German&F1&0.180&0.002&0.395&\textbf{0.626}&0.111\\
&NMI&0.003&\textbf{0.027}&-0.001&0.011&0.020\\
&CA&0.113&0.026&0.443&\textbf{0.622}&0.211\\
\hline
&ARI&0.021&0.063&\textbf{0.409}&0.335&0.267\\
&PS&0.576&0.167&\textbf{0.807}&0.790&0.606\\
Credit&F1&0.233&0.003&\textbf{0.255}&0.169&0.205\\
&NMI&0.013&0.156&\textbf{0.355}&0.329&0.049\\
&CA&0.224&0.018&\textbf{0.266}&0.210&0.032\\
\hline
&ARI&-0.017&0.036&0.000&0.020&\textbf{0.058}\\
&PS&0.751&0.179&0.751&0.756&\textbf{0.797}\\
Adult&F1&0.622&0.040&0.533&\textbf{0.656}&0.009\\
&NMI&0.002&\textbf{0.095}&0.000&0.023&0.055\\
&CA&0.698&0.114&0.500&\textbf{0.756}&0.009\\
\hline
&ARI&0.029&0.004&0.022&0.026&\textbf{0.066}\\
&PS&0.473&0.051&0.441&0.452&\textbf{0.511}\\
CMC&F1&0.294&0.000&\textbf{0.322}&0.299&0.066\\
&NMI&\textbf{0.036}&0.031&0.033&0.030&0.022\\
&CA&0.303&0.005&\textbf{0.324}&0.305&0.009\\
\hline
&ARI&0.085&0.005&0.010&0.000&\textbf{0.151}\\
&PS&0.991&0.590&0.993&0.990&\textbf{1.000}\\
KDD'99&F1&0.001&0.068&\textbf{0.224}&0.000&0.151\\
RIvs.B&NMI&0.069&0.008&\textbf{0.316}&0.000&0.010\\
&CA&0.000&0.187&0.163&\textbf{0.394}&0.028\\
\hline
&ARI&0.092&0.002&0.005&0.092&\textbf{0.113}\\
&PS&0.987&0.590&0.987&0.987&\textbf{0.996}\\
KDD'99&F1&0.001&0.068&\textbf{0.691}&0.007&0.139\\
BOvs.B&NMI&0.031&0.008&\textbf{0.050}&0.031&0.022\\
&CA&0.001&0.187&\textbf{0.583}&0.016&0.139\\
\hline
&ARI&-0.001&0.007&0.008&-0.001&\textbf{0.011}\\
&PS&0.999&0.775&0.999&0.999&\textbf{1.000}\\
KDD'99&F1&0.000&0.024&0.353&\textbf{0.998}&0.011\\
Lvs.S&NMI&0.000&0.007&0.007&0.000&\textbf{0.011}\\
&CA&0.001&0.114&0.236&\textbf{0.998}&0.011\\
\hline
&ARI&-0.001&0.004&\textbf{0.013}&-0.001&0.011\\
&PS&0.998&0.453&0.998&0.998&\textbf{1.000}\\
KDD'99&F1&0.002&0.033&0.556&\textbf{0.997}&0.180\\
Lvs.P&NMI&0.000&\textbf{0.011}&0.010&0.000&0.005\\
&CA&0.002&0.028&0.430&\textbf{0.998}&0.180\\
\hline
&ARI&0.000&-0.003&-0.003&0.000&\textbf{0.118}\\
&PS&0.997&0.776&0.997&0.997&\textbf{0.997}\\
KDD'99&F1&0.000&0.028&\textbf{0.680}&0.000&0.189\\
GPvs.S&NMI&0.000&0.001&0.001&0.000&\textbf{0.003}\\
&CA&0.000&0.113&\textbf{0.585}&0.003&0.102\\
\hline
\end{tabular}
\caption{\label{TableMN}  Results for datasets with mixed attributes. For each case, the winner is highlighted in bold.}
\end{table}
The performance of the weighting schemes for categorical attributes introduced in this work is validated. Introducing two weighting schemes for categorical attributes allows CPF to achieve excellent performance for a broader range of datasets. The KDD '99 datasets contain very imbalanced clusters, with imbalance ratios exceeding 99:1. CPF performs the best on these datasets, indicating that it is more adept than $k$-means-type approaches at detecting clusters with vastly different sizes.

\subsection{Clustering Large Datasets}\label{bd}
The datasets in Table \ref{TableDS2} are used to assess the performance of CPF when clustering large datasets with mixed attributes. The clusterings produced by CPF are compared to those created by the $k$-prototypes method and KAMILA. These are the only comparable algorithms capable of clustering datasets of this size in reasonable time. For both comparison methods, the parameters were specified as suggested and the true number of clusters was provided as an input. As the $k$-prototypes algorithm had not completed running on KDD '99 DOS vs. NORM after 48hrs, it was terminated and no results are presented.

The performance of the CPF method is evident in the results presented in Table \ref{TableBN}. 
\begin{table}[h!]
	\tiny
	\centering
\begin{tabular}{ ccccc}
\hline
& &k-proto.&KAMILA&CPF\\
\hline
&ARI&0.031&0.050&\textbf{0.153}\\
&PS&0.515&0.494&\textbf{0.670}\\
CovType&F1&0.140&0.002&\textbf{0.341}\\
&NMI&0.078&0.132&\textbf{0.202}\\
&CA&0.249&0.063&\textbf{0.341}\\
\hline
&ARI&0.000&0.000&{0.000}\\
&PS&0.501&0.501&{0.501}\\
Poker&F1&0.249&\textbf{0.499}&0.389\\
&NMI&0.000&0.000&{0.000}\\
&CA&0.249&\textbf{0.499}&0.389\\
\hline
&ARI&-&0.000&\textbf{0.089}\\
&PS&-&0.181&\textbf{0.652}\\
KDD'99&F1&-&0.000&\textbf{0.121}\\
DOSvs.NORM&NMI&-&0.000&\textbf{0.020}\\
&CA&-&0.000&\textbf{0.121}\\
\hline
\end{tabular}
\caption{\label{TableBN}  Results for large datasets with mixed attributes. For each case, the winner is highlighted in bold.}
\end{table}
CPF achieves better results across all validity indices for two of the three datasets. Each method struggles to detect meaningful clusters in the Poker dataset. For the CovType dataset, the connected components uncover the underlying structure of the data. For the KDD '99 dataset, the connected component step removes outliers which negatively affect the results of the comparison methods.

\begin{figure}[!h]
	\centering
	\includegraphics[width=\linewidth]{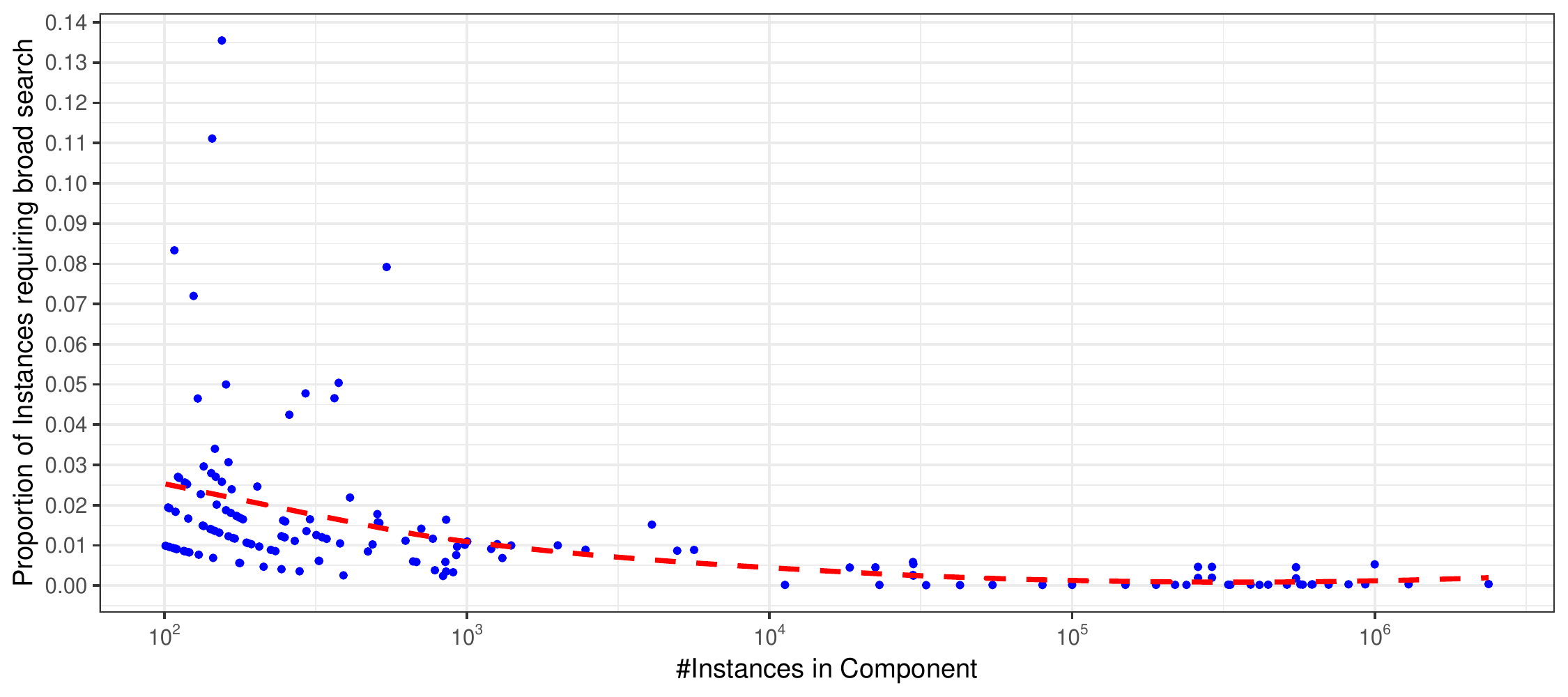}\\
	\caption{Number of instances in a component ($|\mathbb{C}|$) vs. Proportion of instances requiring a broad search ($p$). }
	\label{ps}
\end{figure}

The time taken to cluster each of the datasets is shown in Table \ref{TableTs}. 
\begin{table}[h!]
	\tiny
	\centering
	\begin{tabular}{ cccc}
		\hline
		& $k$-proto. & KAMILA & CPF \\
		\hline
		Cov Type& 	52603.612& 	283.262 & 17782.344\\
		Poker &	74835.257	& 298.656	  & 	30229.664\\
		KDD '99 DOS vs. NORM& -	& 	416.293	& 	137283.899	 \\
		\hline
	\end{tabular}
	\caption{	\label{TableTs} Time (seconds) for each method on large datasets.}
\end{table}
KAMILA is, by some measure, the fastest algorithm for each of the datasets. This reflects the termination rules applied to the $k$-means procedure and its implementation in C++ and Hadoop. The CPF method executes in less time than $k$-prototypes, both implemented in Python. With further optimisation of code, the run time of CPF can be improved. Further context is given to the run time of CPF in Fig. \ref{ps}. 
The proportion of instances which require a broad search, $p$, is shown for a variety of datasets. The observed values of $p$ decrease with the size of the component and is reliably less than 1\% for large components.

\subsection{Parameter Sensitivity Analysis}
\begin{figure}[!h]
	\centering
	\captionsetup[subfigure]{font=footnotesize,labelfont=footnotesize}
	\begin{subfigure}{\linewidth}
		\begin{minipage}{.48\linewidth}
			\centering
			\includegraphics[width=\linewidth]{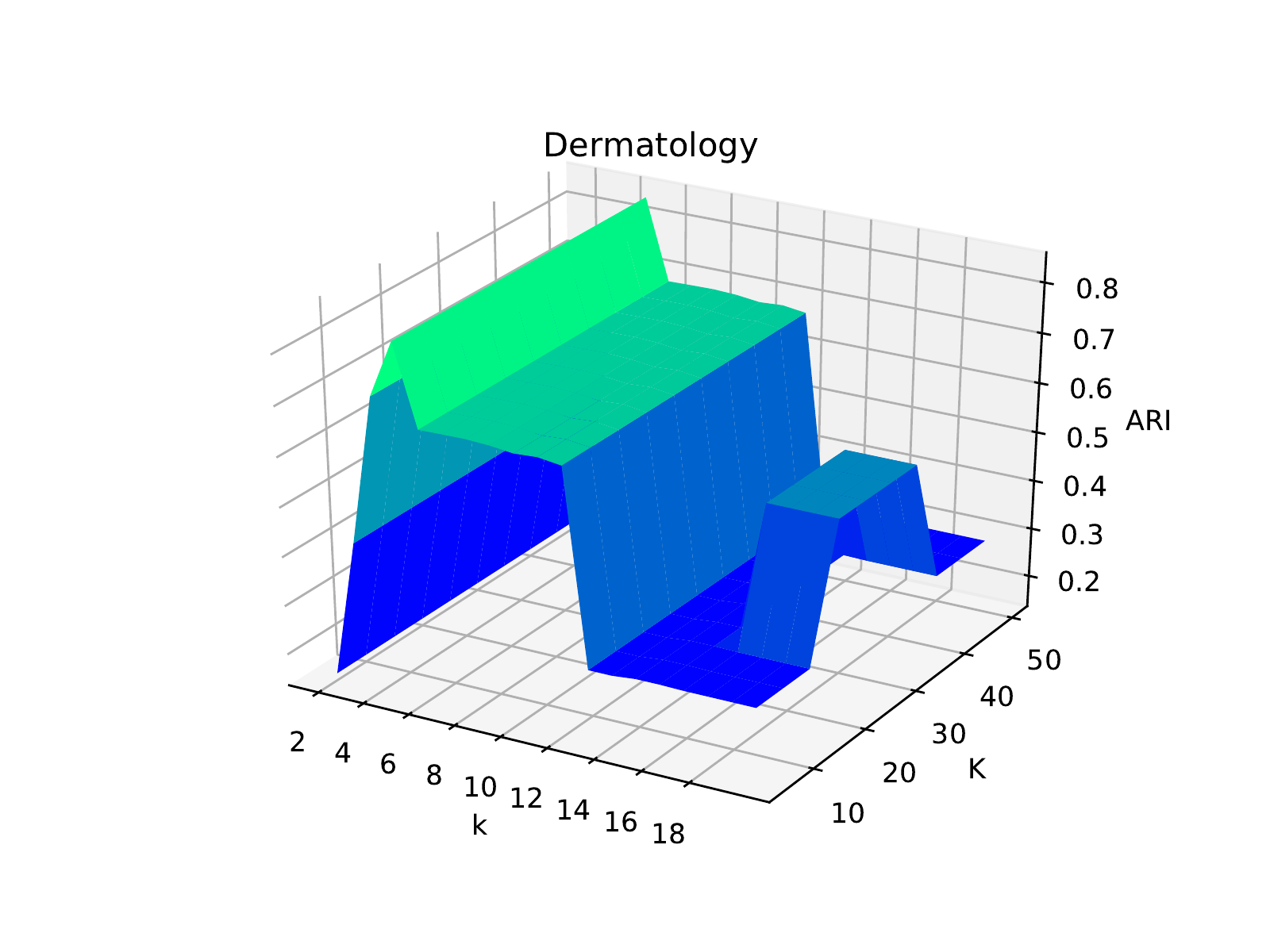}
			
		\end{minipage}%
		\begin{minipage}{.48\linewidth}
			\centering
			\includegraphics[width=\linewidth]{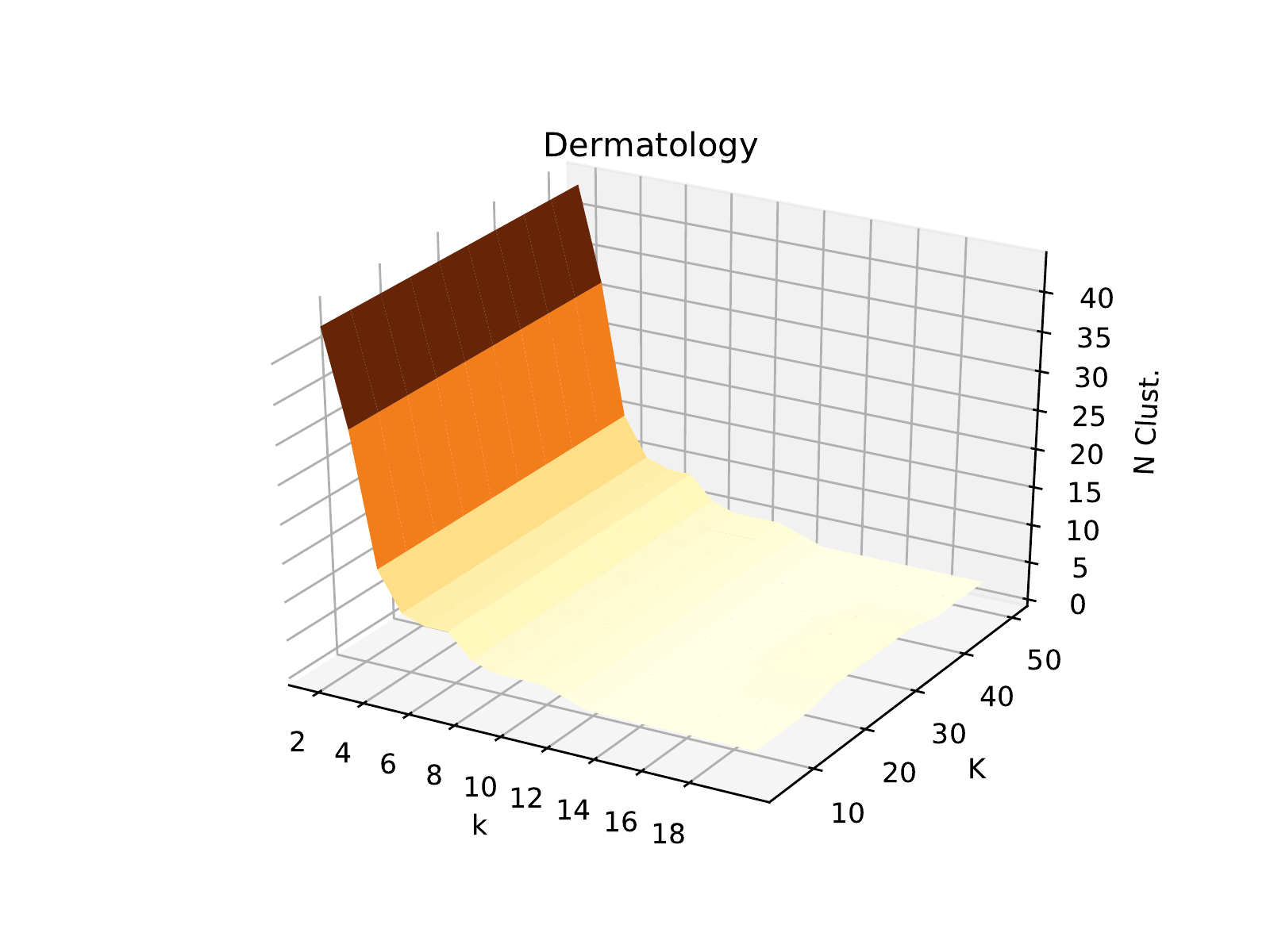}
			
		\end{minipage}
		\caption{Dermatology}
	\end{subfigure}
	\begin{subfigure}{\linewidth}
		\centering
		\begin{minipage}{.48\linewidth}
			\centering
			\includegraphics[width=\linewidth]{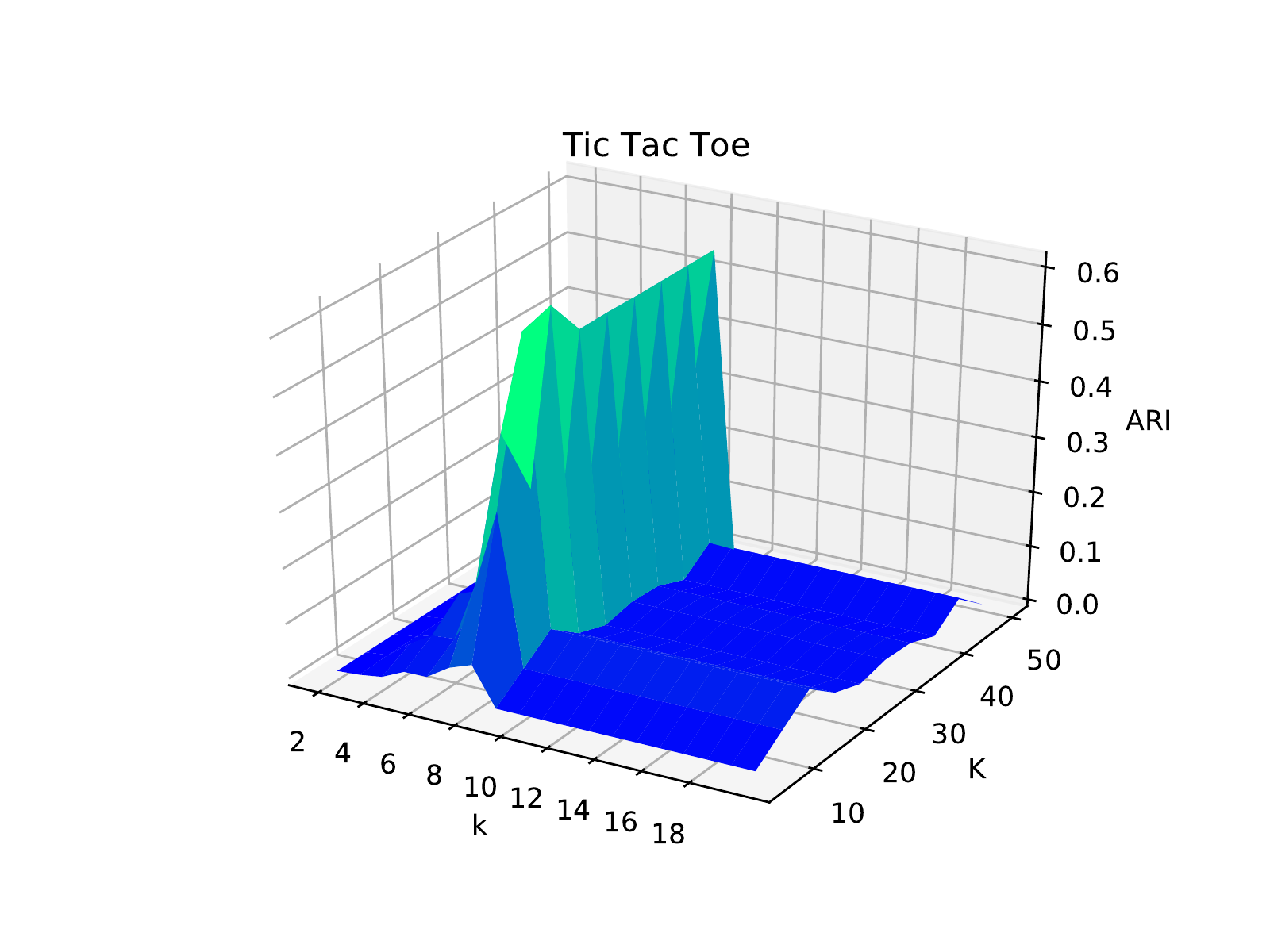}
			
		\end{minipage}%
		\begin{minipage}{.48\linewidth}
			\centering
			\includegraphics[width=\linewidth]{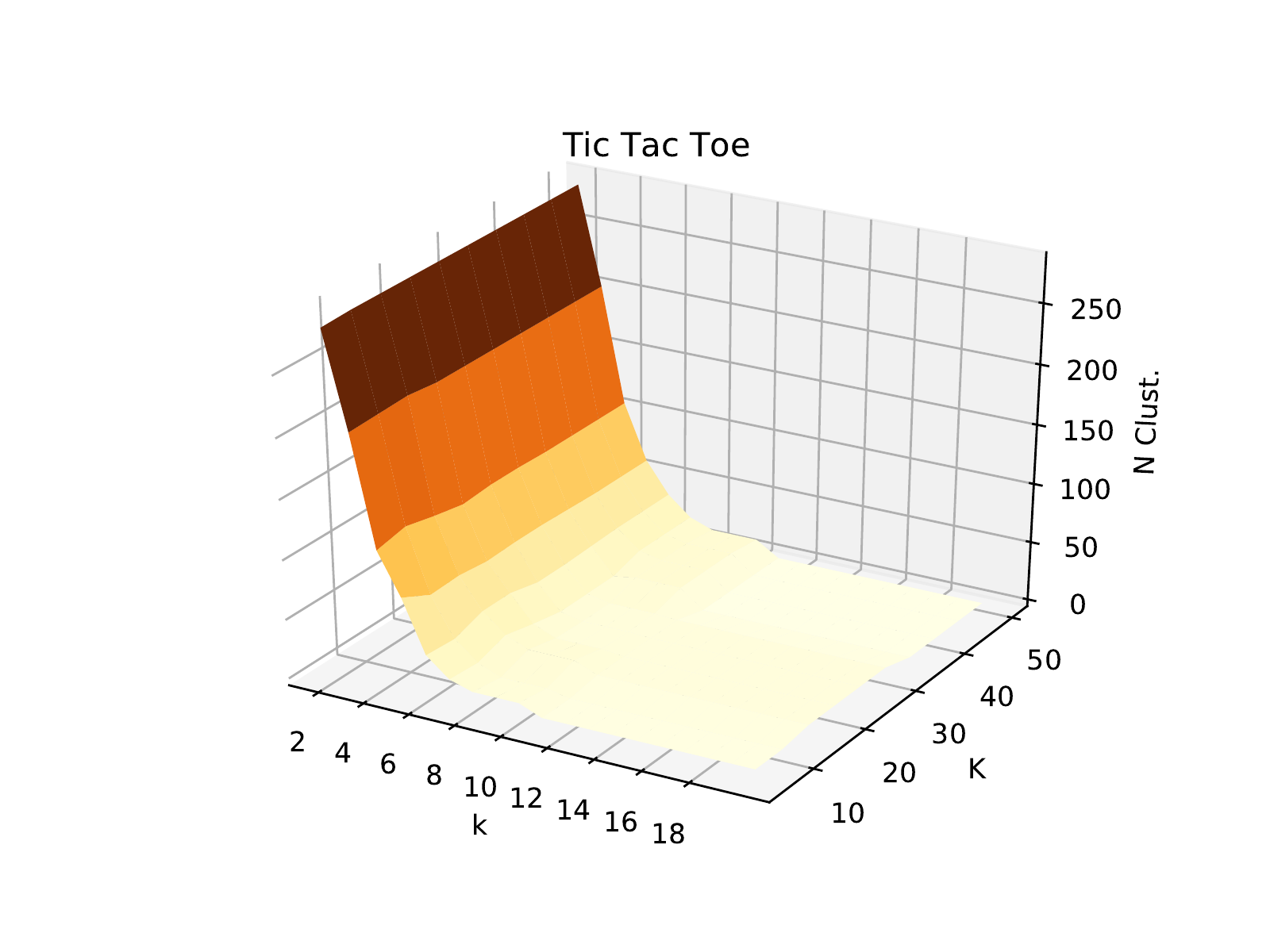}
			
		\end{minipage}
		\caption{Tic Tac Toe}
	\end{subfigure}
	\begin{subfigure}{\linewidth}
		\centering
		\captionsetup[subfigure]{font=footnotesize,labelfont=footnotesize}
		\begin{minipage}{.5\linewidth}
			\centering
			\includegraphics[width=\linewidth]{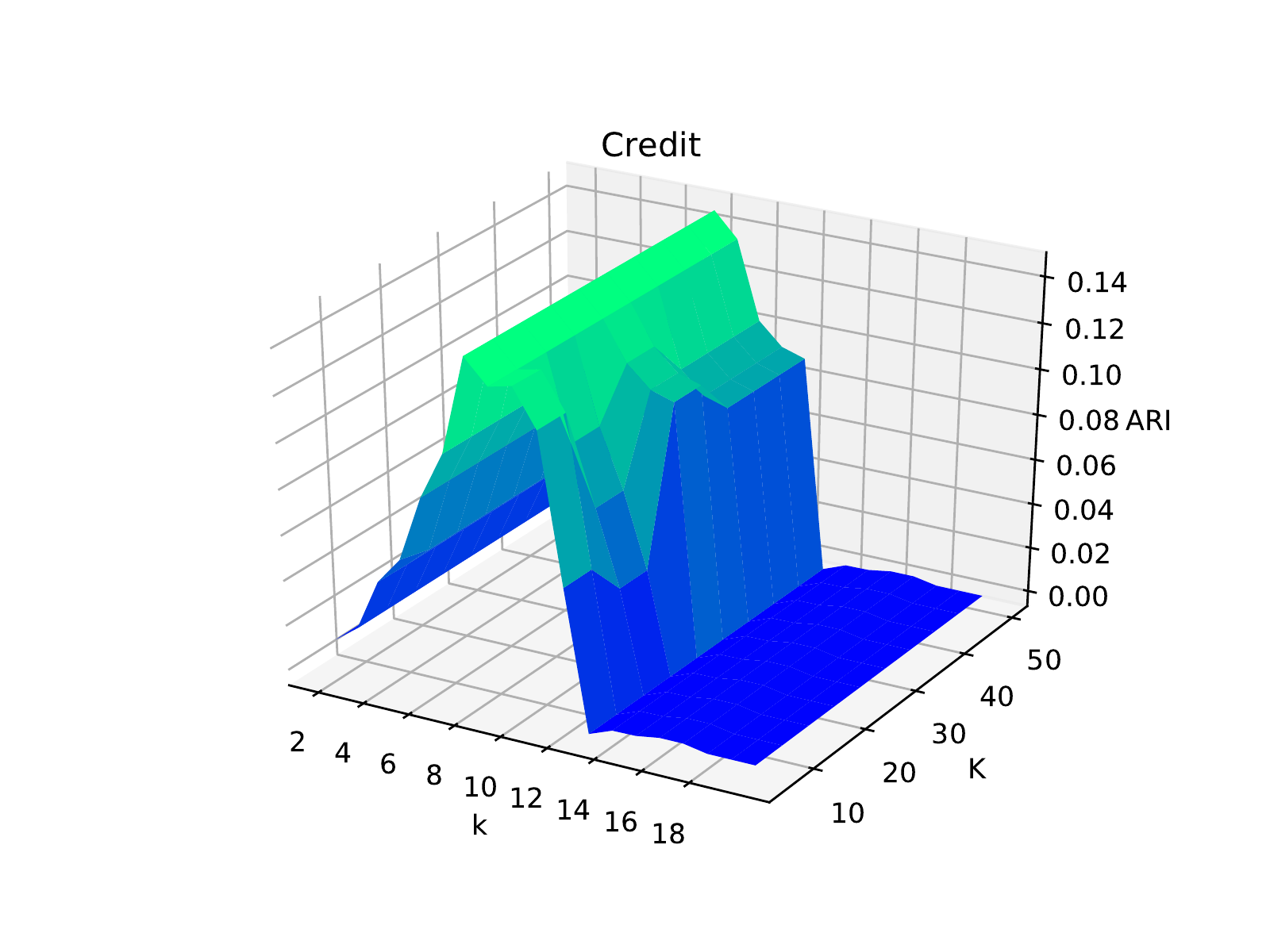}
			
		\end{minipage}%
		\begin{minipage}{.48\linewidth}
			\centering
			\includegraphics[width=\linewidth]{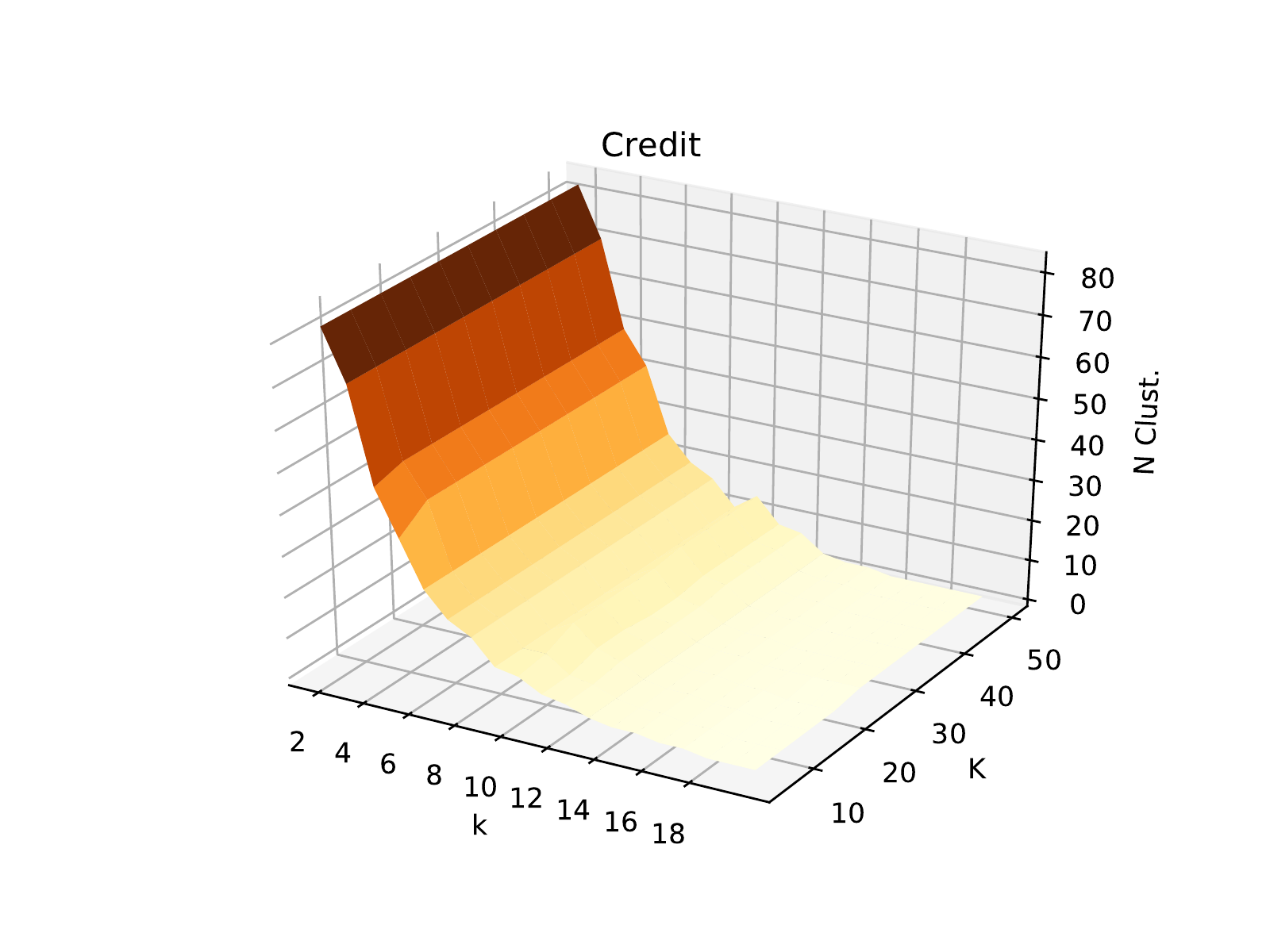}
			
		\end{minipage}
		\caption{Credit}
	\end{subfigure}
	\centering
	\captionsetup[subfigure]{font=footnotesize,labelfont=footnotesize,belowskip=-30pt}
	\caption{\label{params}ARI of CPR vs. $k$ and $K$ (left) and \#Clusters of CPF vs. $k$ and $K$ (right) for Dermatology, Tic Tac Toe, \& Credit datasets.}
\end{figure}

We investigate parameter sensitivity by varying $k$ in $[2, 20]$ and $K$ in $[5, 50]$. We exemplify via three types of datasets: a numerical-attribute dataset, Dermatology, a categorical-attribute dataset, Tic Tac Toe, and a mixed-attribute dataset, Credit. The weighting scheme W1 is used for Tic Tac Toe, while the weighting scheme W2 is used for Credit. We use ARI as a validation index and track the number of clusters resulting from the approach under different input values of $k$ and $K$. Results from the analysis can be seen in Fig. \ref{params}.
\setlength{\belowcaptionskip}{0pt}
It can be observed that the quality of clustering is high when $k$ is within the range $[5, 10]$. Too small a value for $k$ results in connected components which are too fine to capture the underlying cluster structure of the data. If $k$ is too large, the connected components will not capture structures present within the data, by merging regions of the data which are better separated. A small value of $k$ also makes the conductance calculation in the center selection unstable, as the smaller subset can contain very few instances. It can be seen that the choice of $k$ also dominates the number of clusters found in the data.

For the investigated datasets, selecting $K$ in the range $[15, 30]$ yields the best results. This interval contains $\sqrt{n}$ for each of the datasets analyzed. For values in this range, it is less likely that density computation will be influenced by local particularities in the dataset, but not so large as to result in a smoothing of density values, where differences in local density are not captured as the search window is too broad. The results of this analysis echo those reported in Sections \ref{ncm} and \ref{bd} above.

\section{Conclusion \& Future Work}\label{sec5}

This article introduced CPF, an improved algorithm for clustering large mixed-attribute datasets which adapts and extends the peak-finding clustering method. We showed that connected components can be used to achieve better results than peak-finding clustering, while reducing overall computational complexity to $O(n \log n)$. Extensive experimental results further demonstrated that CPF has excellent performance. It achieves superior results over a broad range of benchmark datasets and indices when compared to popular $k$-means-type methods. This performance is driven by the flexibility of CPF, with a novel distance metric and two tuning parameters allowing the user to detect clusters of varying shape, size and density. We leave to future work the investigation of the possibility of applying CPF to data streams and the analysis of network data. Code for the CPF method is available for download from \href{https://pypi.org/project/CPFcluster/}{https://pypi.org/project/CPFcluster/}.

\section*{Acknowledgement}
The authors are grateful to those who provided their source code for use in this work.



\bibliographystyle{apalike}
\bibliography{CPFref}
%

\end{document}